\documentclass[11pt,a4paper]{article}
\usepackage[left=2cm,top=2cm,right=2cm,bottom=1.5cm]{geometry}
\usepackage[small,compact]{titlesec}
\usepackage{url,fancyhdr}
\usepackage{graphicx,rotating,amsfonts,pbox,wrapfig}
\usepackage{amsmath,amsthm,amssymb,color,rotating,enumitem,bibunits}
\usepackage{eurosym,cite}
\usepackage{tikz}
\usetikzlibrary{arrows}

\usepackage{parskip}
\definecolor{yafcolor1}{rgb}{0.4, 0.165, 0.553}
\definecolor{yafcolor2}{rgb}{0.949, 0.482, 0.216}
\definecolor{yafcolor3}{rgb}{0.47, 0.549, 0.306}
\definecolor{yafcolor4}{rgb}{0.925, 0.165, 0.224}
\definecolor{yafcolor5}{rgb}{0.141, 0.345, 0.643}
\definecolor{yafcolor6}{rgb}{0.965, 0.933, 0.267}
\definecolor{yafcolor7}{rgb}{0.627, 0.118, 0.165}
\definecolor{yafcolor8}{rgb}{0.878, 0.475, 0.686}
\definecolor{yafcolor9}{rgb}{0.965, 0.733, 0.767}

\tikzstyle{statnode} = [very thick, draw = yafcolor5, fill = yafcolor5!3, circle, text=black, font=\scriptsize, align = center, rounded corners = 2mm, minimum size = 2.3cm, inner sep = 1pt]
\tikzstyle{statedge} = [thick, >=latex' , ->, yafcolor5]
\tikzstyle{resnode} = [very thick, draw = yafcolor2, rectangle, text=black, font=\scriptsize, align = left, rounded corners = 2mm, dashed, anchor = north]
\tikzstyle{resnodeM} = [ultra thick, draw = yafcolor2, rectangle, text=black, font=\scriptsize, align = left, rounded corners = 2mm, dashed, anchor = north]
\tikzstyle{resedge} = [thick, dashed, yafcolor2,  >=latex' , ->]

\usepackage{url}
\usepackage{authblk}

\begin{document}

\title{Fairness-aware machine learning: a perspective}
\author[1,2]{Indr\.e \v{Z}liobait\.e\thanks{indre.zliobaite@helsinki.fi}}
\affil[1]{Dept. of Computer Science, University of Helsinki, Finland}
\affil[2]{Dept. of Geosciences and Geography, University of Helsinki, Finland}

\maketitle

\begin{abstract}
Algorithms learned from data are increasingly used for deciding many aspects in our life: from movies we see, to prices we pay, or medicine we get. 
Yet there is growing evidence that decision making by inappropriately trained algorithms may unintentionally discriminate people. 
For example, in automated matching of candidate CVs with job descriptions, algorithms may capture and propagate ethnicity related biases. 
Several repairs for selected algorithms have already been proposed, but the underlying mechanisms how such discrimination happens from the computational perspective are not yet scientifically understood.
We need to develop theoretical understanding how algorithms may become discriminatory, and establish fundamental machine learning principles for prevention. We need to analyze machine learning process as a whole to systematically explain the roots of discrimination occurrence, which will allow to devise global machine learning optimization criteria for guaranteed prevention, as opposed to pushing empirical constraints into existing algorithms case-by-case. 
As a result, the state-of-the-art will advance from heuristic repairing, to proactive and theoretically supported prevention. 
This is needed not only because law requires to protect vulnerable people. Penetration of big data initiatives will only increase, and computer science needs to provide solid explanations and accountability to the public, before public concerns lead to unnecessarily restrictive regulations against machine learning. 
\end{abstract}

\section{Background}

This position paper is largely based on my ERC Starting Grant proposal that I wrote in summer of 2015 and slightly updated in summer of 2016. 
The manuscript does not cover related work that came after I submitted the proposal. 
I am sharing a compressed version of the research content not including personal details, descriptions of collaborators, risk management, budget and such. 
My proposal got into the interview round, but did not get funded. A lot of papers came out during the last year and the field is advancing very rapidly, but I think my perspectives outlined here are still up to date and worth sharing.

\subsection{Current situation}

In recent years media and social research have pointed out plenty of anecdotal evidence that decision making by algorithms may unintentionally discriminate people \cite{Datta15,Edelman14,Sweeney13,Kay15,Barocas16,Citron14}. 
News articles raising concerns increasingly appear in the major global press including 
NY Times\footnote{{\scriptsize \url{http://www.nytimes.com/2015/07/10/upshot/when-algorithms-discriminate.html}}},
The Wall Street Journal\footnote{{\scriptsize \url{http://www.wsj.com/articles/computers-are-showing-their-biases-and-tech-firms-are-concerned-1440102894}}}, and
The Guardian\footnote{{\scriptsize \url{http://www.theguardian.com/commentisfree/2015/apr/27/algorithms-are-like-invisible-judges-that-decide-our-fates}}}. 
A few weeks ago the editorial in Nature\cite{Nature16} focused on accountability of big data algorithms.
While attention of the public and media is on gathering evidence, an emerging research discipline of discrimination-aware machine learning and data mining focuses on providing treatment. Existing treatment addresses selected algorithms, but it is not yet clear how to tackle this systematically. 

Discrimination refers to adversary treatment of people based on belonging to some group rather than individual merits. 
For example, automated matching of candidate CVs and job descriptions may propagate ethnicity related biases.
If data encodes human biases, blindly applied machine learning not only picks up and propagates, but possibly even exaggerates those biases \cite{Edelman14}.
More importantly, discrimination by algorithms may occur even if training data is objective, but is not well sampled \cite{Calders13why}.
Human decision makers may occasionally make biased decisions, but 
inappropriately trained models would discriminate continuously and systematically, which may have long term and far reaching effects on large population. 

Discrimination on many grounds, and in many areas of life is forbidden by national and international legislation. 
For instance, in Finland a new act (1325/2014) came into force in January 2015, substantially expanding the scope of protection. 
The act applies to nearly all public and private activities protecting against discrimination based on ethnicity, age, nationality, language, religion, belief, 
opinion, health, disability, sexual orientation or other personal characteristics.
Currently, the EU is preparing a new non-discrimination directive (SEC(2008) 2181), further expanding the scope of protection.
While the scope of protection is expanding worldwide, currently it is not clear how to address potential discrimination by algorithms. 
For example, the US President Office urges to expand technical expertise in preventing disparate impact of big data analytics~\cite{Obama_report}.
A few weeks ago the editorial in Nature highlights a research need to find ways to audit for bias without revealing the algorithms, and to develop new computational techniques that better address and correct discrimination both in training data sets and in the algorithms \cite{Nature16}. 

As the scope and impact of machine learning technologies and decisions informed by big data in the society continues to grow, it is essential to develop scientific expertise for controlling potential discrimination by algorithms. That is not only because the law requires to protect vulnerable people. As computer scientists, we are accountable for algorithm performance, and we need to be able to provide scientific explanations and guarantees of what is happening when algorithms are used for decision support, before public concerns and lack of trust leads to unnecessarily restrictive regulatory actions against machine learning.

\subsection{Research directions for machine learning} 

Computational mechanisms how discrimination by algorithms happens, and when it does not happen, are not yet scientifically well understood. These fundamental insights are needed for developing systematic and theoretically supported prevention. 
A number of approaches have already been proposed suggesting to repair algorithms  by preprocessing training datasets \cite{Feldman15,Mancuhan14,Kamiran13}, adding a regularizer to the model \cite{Calders13,Zemel13,Kamishima12,Kamiran10}, postprocessing trained models \cite{Hajian13}, or model outputs \cite{Kamiran10}. 
However, there are two generic issues with the current approaches. 
Firstly, there is no consensus on how to assess fairness of algorithms, dozens of measures, sometimes contradictory, exist. Typically new solutions come  with new constraints to optimize and measures of performance. 
Without a unifying theory it is hard to argue to what extent current approaches work and generalize. 
Secondly, each solution is typically tailored for a specific combination of machine learning setting and discrimination situation, and hardly generalize to other algorithms, other types of variables, or other grounds of discrimination. 
Proposing tailored repairs for each combination of an algorithm, task setting, and discrimination occurrence 
works in the short run, but this way we are merely treating the symptoms instead of curing the disease.

In the long run we need to develop fundamental scientific understanding for ensuring transparency and accountability of using machine learning in the society. 
Within this goal guaranteeing fairness of algorithms is one of the key issues. 
The research objective is to develop theoretical understanding how algorithms may become discriminatory, and how to prevent it by computational means. Rather than trying to fix case-by-case, my approach is to characterize computational process as a whole in relation to potential unfairness, which would make it possible to uncover the roots of the problem and systematically prevent algorithms from becoming discriminatory. 

My hypothesis, based on our position paper \cite{Calders13why} and a follow up investigation, is that data-driven algorithms become discriminatory due to biased or unrepresentative training data used in combination with global optimization criteria, which focus  on maximizing overall performance not taking into account how the remaining inaccuracies distribute. My expectation is that addressing these issues will not only ensure fairness of algorithms, but also make the overall performance of algorithms better. Currently enforcing fairness is considered as a burden for accurate performance. I believe that addressing the core of the problem will lead to a win-win situation, since guaranteeing fairness would not only be seen as a duty, but will provide clear benefits for everybody involved. 

\subsection{Cross-disciplinary interactions}

\begin{figure}
\centering
\begin{tikzpicture}
\node[statnode, anchor = north] (n1) at (0, 0.5) {\textsc{Computer}\\\textsc{Science}\\[0.3mm]machine learning\\processes\\ 
};
\node[statnode] (n2) at (-3, -4.35) {\textsc{Law}\\[0.3mm]non-\\discrimination\\requirements};
\node[statnode] (n3) at (3, -4.35) {\textsc{Social}\\\textsc{Sciences}\\[0.3mm]fairness\\model};
\node[resnode,align = center] (r2) at (-3, 0.5) {Revising\\non-discrimination\\laws};
\node[resnode,align = center] (r1) at (3, 0.5) {Monitoring\\digital\\discrimination};
\node[resnodeM, align = center] (r3) at (0, -2.8) {{\scriptsize FAIRNESS}\\{\scriptsize AWARE}\\{\scriptsize ALGO-}\\{\scriptsize RITHMS}};
\draw (n1.200) edge[statedge]
    node[sloped, align = center, font=\scriptsize, above, inner sep = 2pt] {expertise about\\[-2pt]algorithms}
    (n2.55);
\draw (n2.30) edge[statedge]
    node[sloped, align = center, font=\scriptsize, below, inner sep = 2pt] {protection\\[-2pt]requirements}
    (n1.225);
\draw (n1.-20) edge[statedge]
    node[sloped, align = center, font=\scriptsize, above, inner sep = 2pt, pos =0.4] {algorithm\\[-2pt]diagnostics}
    (n3.125);
\draw (n3.150) edge[statedge]
    node[sloped, align = center, font=\scriptsize, below, inner sep = 2pt] {what behavior\\[-2pt]is fair} 
    (n1.-45);
\draw (n1.32) edge[resedge] (r1);
\draw (n3) edge[resedge] (r1);
\draw (n1.148) edge[resedge] (r2);
\draw (n2) edge[resedge] (r2);
\draw (n1) edge[resedge,ultra thick] (r3);
\draw (n2.-10) edge[very thick,dotted, <->, >=latex', gray] (n3.190);
\end{tikzpicture}
\caption{Cross-disciplinary interactions.}
\label{fig:multi}
\end{figure}
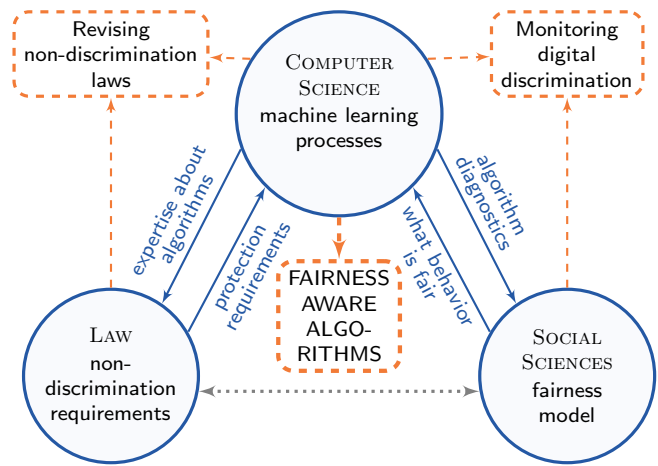

The research is at an intersection of computer science, law and social sciences, with the main focus in machine learning.
Interaction with social sciences helps to define fair resource allocation for diagnosing; computer science is responsible for analysis of machine learning processes, computing methodologies, and performance guarantees; interaction with law helps to prescribe non-discrimination requirements, and computationally guarantee prevention. Figure \ref{fig:multi} depicts the setting, solid lines show cross-disciplinary interactions, and dashed lines show what would become possible. 

The aims for machine learning research is to provide computational means to account for potential discrimination. 
We do not intend to judge what behavior is fair or not, or reveal any concrete cases of discrimination. 
From the machine learning research perspective it is not essential on which grounds (e.g. race, age, gender), or in which areas (e.g. work, education, housing) potential discrimination is. Yet the expertise of collaborators in law and social science is essential for establishing general principles for diagnosing and preventing. 

\section{Proposed research} 

\subsection{State-of-the-art (as of summer 2016)} 

Discrimination-aware machine learning and data mining is an emerging discipline studying potential discrimination by algorithms, transparency and accountability of using algorithms for decision support. The discipline builds upon machine learning and data mining in algorithm design, statistical science in discrimination testing, as well as sociology and law in defining fairness.
The goal so far has been to develop algorithmic techniques that would obey fairness regulations prescribed by law.
The focus has been on two main challenges: how to formulate fairness constraints mathematically, and how to make models to obey them.
An interdisciplinary survey~\cite{Romei14} overviews techniques used, a legal review \cite{Barocas16} presents legal background. 
While attention of the public and media is on gathering evidence, the focus of the emerging research community is on providing treatment. 
Yet, the mechanisms of \emph{how} discrimination by algorithms happens, \emph{in which circumstances} it happens, and \emph{when it does not happen} are so far not well understood. 

Existing research follows two directions: discrimination discovery, and discrimination prevention. 
Discovery aims at finding discriminatory patterns in data using data mining and machine learning methods \cite{Ruggieri10,Mancuhan14,Luong11}.
It builds upon extensive research in statistics on discrimination evidence (e.g. \cite{Kaye82,Tinkham10,Blank04}), 
addressing new challenges due to increasing volumes and complexity of data and ways of possible unfairness.
Statistics has been focusing on hypotheses testing in decision data, and provide essential solutions how to correctly compare groups of people. 
Prevention is an orthogonal issue focusing on machine learning processes, which is becoming increasingly urgent due to omnipresent big data initiatives.
Discrimination prevention develops methods for sanitizing algorithms or adjusting machine learning process such that outputs would obey selected fairness constraints.  
Several attempts to fix algorithms have been made by preprocessing training datasets \cite{Feldman15,Mancuhan14,Kamiran13}, adding a regularizer to the model \cite{Calders13,Zemel13,Kamishima12,Kamiran10}, postprocessing trained models \cite{Hajian13} or model outputs \cite{Kamiran10}. 

However, there are two generic issues with the current approaches. 
Firstly, there is no consensus on how to measure fairness of algorithms (dozens of measures, sometimes contradictory, exist), therefore, it is hard to argue to what extent the proposed fixes work and generalize. 
Secondly, each solution is typically tailored for a specific setting and situation, and it is hard to generalize to other algorithms, other types of variables, or other grounds of discrimination. 
Proposing tailored repairs for each combination 
is like coping with symptoms instead of curing the disease.

I propose instead of repairing algorithm by algorithm in variants of settings to characterize computational process as a whole in relation to potential unfairness. This can be achieved by analyzing interactions between data characteristics, learning task settings in relation to protected variables, and learning techniques with their assumptions. Identifying \emph{which} assumption violations are critical, and theoretically quantifying the effects of those violations, taking into account special status of (potentially many) protected variables, to be formalized in collaboration with social science and law, promise technical contributions in computer science, not only for solving the cross-disciplinary problem, which is the main goal, but also for better understanding the implications of imperfect data for machine learning in general.  
The main technical research challenge is to derive expectations and guarantees of performance in terms of thoroughly defined fairness measures.

Such an approach would lead to theory of potential discrimination occurrence in machine learning, backed up with expectations for unfairness, and accompanied with principles for prevention. 
It will become possible to \emph{anticipate unfairness} and prevent it systematically even before the initial models are built, as opposed to repairing models when discrimination is discovered. 

\subsection{Research tasks}

The research objective is to develop theoretical understanding how algorithms may become discriminatory, and establish fundamental machine learning principles for prevention. 
Machine learning research is not aiming at pointing out any particular discrimination cases, or judging what behavior is right or wrong. 
The goal is to develop technological knowledge. This can be reached by solving the following research tasks:

\begin{itemize}
\item \textbf{RQ1} [diagnosing] Consolidated measures of fairness of algorithms;
\item \textbf{RQ2} [explaining] Empirical and theoretical analysis how algorithms become discriminatory;
\item \textbf{RQ3} [preventing] Fairness-aware optimization process for machine learning models.
\end{itemize}

Solving RQ1 enables systematic diagnosing, RQ2 develops theory for on understanding of the phenomenon, and RQ3 develops mechanisms for making models fair in such a way that not only the fairness constraints are satisfied, but models become better. 
For RQ1 I have already obtained preliminary results \cite{Zliobaite15,Zliobaite15Arxiv} (showing the potential of ROC), 
and have a very concrete action plan. The uncertainty for RQ1 is low and the roadmap how to proceed is clear. 
This also makes a good starting point for a PhD student.
RQ2 is set to investigate hypotheses outlined in our position paper \cite{Calders13why}, and potentially generate new hypotheses. 
As a plan minimum we can work through our existing hypotheses and provide theoretical evidence for those, which will already make a major contribution. 
Optimization criteria in RQ3 depend on theoretical understanding gained in RQ2. 
I anticipate to go beyond the current understanding that enforcing fairness is at a cost of accuracy. 
Via redesigning machine learning optimization criteria I hope to demonstrate that enforcing fairness also leads to more accurate models. 
Even if this does not succeed to the maximum, devising global optimization criteria to guarantee fairness and minimize costs in accuracy will be a notable contribution, in addition to contributions in RQ1 and RQ2.

The main research output will be a methodological framework, translating types of task settings, base models, modeling assumptions, and data characteristics into distinct schemes of discrimination occurrence, backed up with theoretical expectations of discrimination, accompanied with principles for prevention. Given these foundations, computer scientists can develop new algorithmic techniques for prevention, law experts can investigate how to reflect these technical findings in non-discrimination laws, sociologists can develop new analysis tools, general public can have guarantees of non-discrimination.

My working hypothesis is that algorithms become discriminatory due to specific violations of assumptions upon which learning methods are usually based, 
in combination with optimization criteria used in learning. 
More specifically, discrimination may occur when training data is incorrect (e.g. contain discriminatory decisions in the past), incomplete (e.g. omitted variable bias), or badly sampled (e.g. some groups are over/underrepresented, or representation changes over time) \cite{Calders13why}. 
Discrimination by algorithms may occur even if there is no unfairness in the training data. 
A typical machine learning objective is to maximize global performance, not controlling how the remaining inaccuracies distribute across groups of individuals.
My hypothesis for prevention is that rather than forcing non-discrimination constraints algorithm by algorithm, we can use a generic robust optimization criteria, maybe sacrificing a bit of overall accuracy, but minimizing large one-directional errors within subgroups.

\section{Proposed research directions}

\subsection{Consolidated measures of fairness of algorithms}

The objective of RQ1 is to scientifically justify the performance criteria for fairness in decision making by algorithms. 
Literally, dozens of discrimination measures have been used (many, but not all are listed in \cite{Romei14}), differing in input variables, use of statistical tests, or comparison functions, measuring the share of people discriminated or the magnitude of discrimination, counting decisions from data or inferring models and inspecting their parameters, taking or not taking into account any factors explaining the difference between people. 
Measures for direct discrimination (identify discriminated people), and indirect discrimination (assess fairness of decision rules) are sometimes mixed together. 
As the field is emerging, researchers in exploratory phase come up with new measures and definitions. 
There is a lack of scientific arguments for using one or the other, and it is not clear what researchers are after. 

Our recent analysis \cite{Zliobaite15,Zliobaite15Arxiv} suggests that instead of searching for discrimination in decision data and trying to remove it, we need to consider what makes algorithms discriminatory. 
As an example, consider automated CV and job matching service, where an algorithm is used to decide, who gets invited for an interview. 
The algorithm ranks all CVs using their text as input, and decision makers decide how many of the top-ranked candidates to invite.
If in the rank list all the nationality X candidates appear before all nationality Y candidates, 
then the \emph{algorithm} is maximally biased in favor of X\footnote{Assuming that the distributions of qualifications of X and Y are equal, for illustration assume they are. If not, then there are good solutions from statistics to handle that, e.g. propensity score matching \cite{Pearl09}.}. 
However, suppose there is enough capacity, and all the candidates are invited for an interview, then the \emph{decision data} shows no discrimination. 
But the algorithm still is biased to the maximum, and may discriminate in the next application round.
Researchers have been using measures built upon indirect discrimination statistics (e.g. inspired by the t-test).
My insights suggest a conceptually different approach to look into the ranking mechanism, because that is due to algorithm.

We can start from defining a mathematical model for fairness (RQ1.1), consulting social scientists, and experts in net-neutrality, and describe discrimination mechanisms by equations (similarly to what we did in \cite{Zliobaite11icdm} but at a larger scope).
Then we can analytically and experimentally evaluate existing measures (RQ1.2) and select a principle that capture all aspects of the fairness model, including accounting for explanatory variables (RQ1.3).
Our recent investigation suggests that ROC curve alike measures could be suitable. 
Similar measures have episodically been used as an alternative for measuring discrimination in a regression context \cite{Calders13}. 
ROC suffers from computational complexity (RQ1.4) when working with large datasets, but there are workarounds, e.g. rank approximations. 
Perhaps building upon recent studies on ROC curves in cost space \cite{Hernandez13} will help to establish the final arguments. 

\subsection{Empirical and theoretical analysis how algorithms become discriminatory}

Solving RQ1 task will enable us to characterize performance of algorithms in terms discrimination. When we are able to measure discrimination, and have scientific confidence that what we are measuring is what we are after, we can proceed to the core of this research RQ2, that is, analyze in which circumstances algorithms inferred from data become discriminatory, and to what extent. We have several hypotheses and preliminary evidence outlined in our position paper \cite{Calders13why}, namely, that discrimination happens when training data is incorrect (biased) (RQ2.1), incomplete in feature space (RQ2.2), or in the instance space (RQ2.3). 
We should also consider evolving data (RQ2.4).
We can analyze in-depth these scenarios on synthetic and real benchmark datasets (such as census income) on a full range of task settings: classification and regression, binary, numeric, categorical protected ground, multiple protected grounds, categorical and numerical explanatory variables. 
On synthetic data we can control the magnitude of biases and degree of data incompleteness (w.r.t. to the assumed data generation process) to fine tune the hypotheses, for which we can seek theoretical explanations in a similar spirit similarly as, for example, in missing value analysis for streaming data \cite{Zliobaite15ML}. 
Some explanations, such as the so called Simpson's paradox (aggregated data may exhibit reverse patterns than subgroups), or the omitted variable bias, will be based on causality theory \cite{Pearl09}. 
Since currently thorough explanations are lacking, any explanatory theories here will be novel and make fundamental contributions.

RQ2 should also revisit the open question of keeping or not keeping the protected characteristic (e.g.~age) in the model. 
It has been clearly demonstrated that removing the protected characteristic does not solve the discrimination problem (e.g. \cite{Kamiran10}),
yet lawyers are strictly for removing it anyway (personal communication).   
My pilot investigation \cite{Zliobaite15} suggests that in the model training process it has to stay for being able to quantify, which portion of observed inequality is justifiable, and which should be eliminated. 
Final models for decision making should not use it, of course, but it seems that the initial machine learning has to. 
RQ2 will study this systematically, and provide scientific arguments, which should provide direct implications to policy making. 
The current tendency is to forbid collecting any sensitive data, but it is very likely that unless it is allowed, algorithm developers would not be able to guarantee non-discrimination.
Our recent pilot study with linear regression \cite{Zliobaite16AI} demonstrates how and why this can happen. We present a toy example and a theoretical explanation in the Appendix. This is intended to exemplify how in principle this kind of research can be done. 

Following this analysis a methodology for auditing algorithms will be developed in RQ2.5. It will include diagnostics framework from RQ1, and a procedural check-list for model development (based on what may go wrong) from RQ2.1-RQ2.4.

Solving RQ2 will equip us with theoretical understanding on how discrimination by algorithms happens, and will form fundamentals for discrimination-aware machine learning.
This will make it possible to:
(1) diagnose existing algorithms (e.g. a bank wants to assess their credit scoring model in terms of potential discrimination),
(2) judge intended machine learning procedures (e.g. a bank is going to make a new credit scoring model and wants to assess the procedure they are going to use), 
(3) updating existing, and developing new machine learning techniques (e.g. a researcher develops a new deep learning technique) that would not only inherently prevent discrimination, but also guarantee discrimination prevention.

\subsection{Fairness-aware optimization criteria for machine learning}

Building on RQ2 we will tackle RQ3 to come up with novel optimization criteria for machine learning when used in decision support. 
Standard machine learning one way or the other optimizes overall performance accuracy. 
In discrimination-aware machine learning variants are tested against non-discrimination constraints \cite{Feldman15,Mancuhan14}, or models are regularized using constraints \cite{Calders13,Zemel13,Kamishima12,Kamiran10}.
This works in a particular scenario on a particular test set with particular constraints, but it is not realistic to consider all possible constraints for all possible protected grounds, and data may vary over time as well. 

Anticipating the solution now is a bit tricky, since it will depend on findings in RQ2. 
A remotely related attempt \cite{Calders13} does not guarantee non-discrimination with the chosen measure of success.
My scientific intuition is that the optimization criteria should be robust in a way to minimize large deviations of errors across subgroups in the instance space (RQ3.2). 
One approach would be to cluster inputs (or apply propensity score matching \cite{Pearl09}) and then optimize accuracy per cluster instead of per observation, at the same time controlling the direction of deviations. I also expect that something similar to what we have developed for regression with missing data \cite{Zliobaite15ML} could work, where a parameter controls the tradeoff between ideal accuracy and robustness. 
In case we cannot come up with a criterion based solely on accuracy (target and prediction), we will add non-discrimination constraints including the protected variable. 

In addition, RQ3 will develop better understanding of tradeoffs between accuracy and discrimination (RQ3.1). 
So far it has been commonly accepted that accuracy may decrease \cite{Kamiran10,Zliobaite15}, but there are many covenants. 
If testing data is discriminatory, then the observed testing accuracy is flawed.
The ultimate goal in RQ3 is by rethinking the optimization criteria to demonstrate that enforcing non-discrimination also leads to better models. 
If successful, not only this will change how researchers think about machine learning process, but provide sound scientific arguments beyond "the law requires so" to adopt non-discriminatory algorithms in practice (RQ3.3).

Discrimination-aware machine learning and data mining is an emerging discipline, researchers are exploring possibilities, new solutions keep revealing new challenges, and research has only scratched the surface so far. 
In this situation all the outlined research tasks, anticipating a systematic approach to the problem and a unifying theory, have a good potential for scientific breakthroughs. And the highest potential for scientific breakthroughs is perhaps in RQ3, which may make to rethink overall performance and optimization criteria for building machine learning that are intended to be used for decision support.

\bibliographystyle{plain}
\bibliography{bib_discrimination2}

\begin{thebibliography}{10}

\bibitem{Barocas16}
S.~Barocas and A.~D. Selbst.
\newblock Big data's disparate impact.
\newblock {\em California Law Review}, 104, 2016.

\bibitem{Blank04}
R.~M. Blank, M.~Dabady, C.~F. Citro, and National Research Council (U.S.)~Panel
  on~Methods~for Assessing~Discrimination.
\newblock {\em Measuring racial discrimination}.
\newblock National Academies Press, 2004.

\bibitem{Calders13}
T.~Calders, A.~Karim, F.~Kamiran, W.~Ali, and X.~Zhang.
\newblock Controlling attribute effect in linear regression.
\newblock In {\em Proc. of 13th IEEE ICDM}, pages 71--80, 2013.

\bibitem{Calders13why}
T.~Calders and I.~Zliobaite.
\newblock Why unbiased computational processes can lead to discriminative
  decision procedures.
\newblock In {\em Discrimination and Privacy in the Inf. Society}, pages
  43--57. 2013.

\bibitem{Citron14}
D.~K. Citron and F.~A. Pasquale.
\newblock The scored society: Due process for automated predictions.
\newblock {\em Washington Law Review}, 89(1), 2014.

\bibitem{Datta15}
A.~Datta, M.~C. Tschantz, and A.~Datta.
\newblock Automated experiments on ad privacy settings a tale of opacity,
  choice, and discrimination.
\newblock {\em Proc. on Privacy Enhancing Technologies}, 2015(1):92--112, 2015.

\bibitem{Edelman14}
B.~G. Edelman and M.~Luca.
\newblock Digital discrimination: The case of {Airbnb.com}.
\newblock Working Paper 14-054, Harvard Business School, 2014.

\bibitem{Nature16}
Editorial.
\newblock More accountability for big-data algorithms.
\newblock {\em Nature}, 537(7621), 2016.

\bibitem{Feldman15}
M.~Feldman, S.~A. Friedler, J.~Moeller, C.~Scheidegger, and
  S.~Venkatasubramanian.
\newblock Certifying and removing disparate impact.
\newblock In {\em Proc. of 21st ACM KDD}, pages 259--268, 2015.

\bibitem{Hajian13}
S.~Hajian and J.~Domingo{-}Ferrer.
\newblock A methodology for direct and indirect discrimination prevention in
  data mining.
\newblock {\em {IEEE} Trans. Knowl. Data Eng.}, 25(7):1445--1459, 2013.

\bibitem{Hernandez13}
J.~Hernandez-Orallo, P.~Flach, and C.~Ferri.
\newblock {ROC} curves in cost space.
\newblock {\em Machine Learning}, 93(1):71--91, 2013.

\bibitem{Obama_report}
The~White House.
\newblock {\em Big Data: Seizing Opportunities, Preserving Values}.
\newblock 2014.

\bibitem{Kamiran10}
F.~Kamiran, T.~Calders, and M.~Pechenizkiy.
\newblock Discrimination aware decision tree learning.
\newblock In {\em Proc. of 10th IEEE ICDM}, pages 869--874, 2010.

\bibitem{Kamiran13}
F.~Kamiran, I.~Zliobaite, and T.~Calders.
\newblock Quantifying explainable discrimination and removing illegal
  discrimination in automated decision making.
\newblock {\em Knowl. Inf. Syst.}, 35(3):613--644, 2013.

\bibitem{Kamishima12}
T.~Kamishima, S.~Akaho, H.~Asoh, and J.~Sakuma.
\newblock Fairness-aware classifier with prejudice remover regularizer.
\newblock In {\em Proc. of ECMLPKDD}, pages 35--50, 2012.

\bibitem{Kay15}
M.~Kay, C.~Matuszek, and S.~Munson.
\newblock Unequal representation and gender stereotypes in image search results
  for occupations.
\newblock In {\em Proc. of 33rd ACM CHI}, pages 3819--3828, 2015.

\bibitem{Kaye82}
D.~Kaye.
\newblock Statistical evidence of discrimination.
\newblock {\em Journal of the American Statistical Association}, 77(380), 1982.

\bibitem{Luong11}
B.~T. Luong, S.~Ruggieri, and F.~Turini.
\newblock {k-NN} as an implementation of situation testing for discrimination
  discovery and prevention.
\newblock In {\em Proc. of 17th KDD}, pages 502--510, 2011.

\bibitem{Mancuhan14}
K.~Mancuhan and C.~Clifton.
\newblock Combating discrimination using bayesian networks.
\newblock {\em Artif. Intell. Law}, 22(2):211--238, 2014.

\bibitem{Pearl09}
J.~Pearl.
\newblock {\em Causality: Models, Reasoning and Inference}.
\newblock Cambridge University Press, 2009.

\bibitem{Romei14}
A.~Romei and S.~Ruggieri.
\newblock A multidisciplinary survey on discrimination analysis.
\newblock {\em Knowledge Eng. Review}, 29(5):582--638, 2014.

\bibitem{Ruggieri10}
S.~Ruggieri, D.~Pedreschi, and F.~Turini.
\newblock Data mining for discrimination discovery.
\newblock {\em ACM Trans. Knowl. Discov. Data}, 4(2):9:1--9:40, May 2010.

\bibitem{Sweeney13}
L.~Sweeney.
\newblock Discrimination in online ad delivery.
\newblock {\em Commun. of the ACM}, 56(5):44--54, 2013.

\bibitem{Tinkham10}
T.~Tinkham.
\newblock The uses and misuses of statistical proof in age discrimination
  claims.
\newblock {\em Hofstra Labor and Emploment Law Journal}, 27, 2010.

\bibitem{Zemel13}
R.~S. Zemel, Y.~Wu, K.~Swersky, T.~Pitassi, and C.~Dwork.
\newblock Learning fair representations.
\newblock In {\em Proc. of 30th ICML}, pages 325--333, 2013.

\bibitem{Zliobaite15}
I.~Zliobaite.
\newblock On the relation between accuracy and fairness in binary
  classification.
\newblock In {\em Proc of the 2nd workshop on Fairness, Accountability, and
  Transparency in Machine Learning}, 2015.

\bibitem{Zliobaite15Arxiv}
I.~Zliobaite.
\newblock A survey on measuring indirect discrimination in machine learning.
\newblock {\em CoRR}, abs/1511.00148, 2015.

\bibitem{Zliobaite16AI}
I.~Zliobaite and B.~Custers.
\newblock Using sensitive personal data may be necessary for avoiding
  discrimination in data-driven decision models.
\newblock {\em Artif. Intell. Law}, 24(2):183--201, 2016.

\bibitem{Zliobaite15ML}
I.~Zliobaite and J.~Hollmen.
\newblock Optimizing regression models for data streams with missing values.
\newblock {\em Machine Learning}, 99(1):47--73, 2015.

\bibitem{Zliobaite11icdm}
I.~Zliobaite, F.~Kamiran, and T.~Calders.
\newblock Handling conditional discrimination.
\newblock In {\em Proc. of 11th IEEE ICDM}, pages 992 -- 1001, 2011.

\end{thebibliography}

\newpage

\section*{Appendix: How linear regression models may end up with algorithmic discrimination}

We demonstrate empirically and theoretically with standard regression models  \cite{Zliobaite16AI} that in order to make sure that decision models are non-discriminatory, for instance, with respect to race, the sensitive racial information needs to be used in the model building process. Of course, after the model is ready, race should not be required as an input variable for decision making. From the regulatory perspective this has an important implication: collecting sensitive personal data is necessary in order to guarantee fairness of algorithms, and law making needs to find sensible ways to allow using such data in the modeling process.

Imagine a simplified society, where monthly salary is assumed to depend on years of education, but in fact, due to some prejudice, it actually depends not only on education, but also on ethnicity of a person (e.g. native or immigrant). 
Suppose that the true underlying mechanism how salaries are decided is:
\begin{equation}
\mathit{salary} = 1000 + 100 \times \mathit{education} - 500 \times \mathit{ethnicity}, 
\label{eq:toy}
\end{equation}
where $\mathit{education}$ represents years of education, $\mathit{ethnicity}$ is $0$ for natives and $1$ for immigrants. 
That is, people with no education get $1000$ base salary, and for every year of education there is $100$ extra. Immigrants get $500$ less in all circumstances. This is a fictitious example. 

Consider that the decision rules are unknown to the society, 
and data a data scientist wants to develop a salary recommendation system using observed data, given in Table \ref{tab:toy}.
Either due to belief that salary should depend only on education, but not ethnicity, or due to legislation explicitly forbidding to use ethnicity, the data scientist decides to omit ethnicity. The assumed model form is
$\mathit{salary} = b_0 + b_1 \times \mathit{education},$ 
where $b_0$ is a coefficient denoting the base salary, and $b_1$ is a coefficient denoting how much extra a person should get for each extra year of education. The data scientist will find the coefficients from the data. 
\begin{table}[h]
\centering
\caption{Toy example: synthetic data about salaries, generated according to Eq.~(\ref{eq:toy}).}
\begin{tabular}{ccccccc}
\hline
$\mathit{education}$ &  $\mathit{ethnicity}$ & $\mathit{salary}$ & & $\mathit{education}$ &  $\mathit{ethnicity}$ & $\mathit{salary}$\\
\hline
1 & 1 & 600 &   & 1 & 0 & 1100 \\ 
2 & 1 & 700 &   & 6 & 0 & 1600 \\ 
3 & 1 & 800 &   & 7 & 0 & 1700 \\ 
4 & 1 & 900 &   & 9 & 0 & 1900 \\ 
10 & 1 & 1500 &   & 10 & 0 & 2000 \\ 
\hline
\end{tabular}
\label{tab:toy}
\end{table}

After running the standard regression fitting procedure (ordinary least squares) on the data in Table \ref{tab:toy}, the following model is obtained:
\begin{equation}
\mathit{salary} = 602 + 128\times \mathit{education}. 
\end{equation}
We can see that people with no education would get $602$ base salary, instead of $1000$ which they are supposed to get according to the ground truth in Eq.~(\ref{eq:toy}). In addition, for every extra year of education a person would get $128$, which is more than it is supposed to be according to the true underlying process ($100$). The fitted model punishes people with low education more than necessary, and rewards people with high education more than deserved, which is by itself already incorrect reflection of the underlying process, and would make incorrect recommendations, if used as a recommender system for salaries. 

Moreover, in this fictitious society immigrants tend to have lower education, as it can be seen from Table~\ref{tab:toy}. The education variable is correlated with ethnicity. Hence, not only the learned model is incorrect, but immigrants suffer from that incorrectness more, because they tend to have lower education. 

The example demonstrates that removing the ethnicity from the modeling process does not ensure that the model is free from discrimination. If instead the data scientist includes the sensitive attribute when fitting the model, and afterwards removes \emph{the model component} related to ethnicity, the resulting model will be correct in terms of education and free from discrimination with respect to ethnicity. 

Fitting a regression model on the complete data in Table \ref{tab:toy} recovers $\mathit{salary} = 1000 + 100 \times \mathit{education} - 500 \times \mathit{ethnicity}$. Now we can remove the ethnicity component, replacing it by a constant which does not depend on ethnicity, in order to get a correct and discrimination-free model for recommendations: 
\begin{equation}
\mathit{salary} = 1000 + 100 \times \mathit{education} - c.
\end{equation}
The constant $c$ can be zero, or computed from the data as will be discussed in the next section. 
We can argue that as long as $c$ is the same for both ethnicities, this model will not discriminate, because the salary will correctly reflect the base level, and the variable component for extra education (as per underlying Eq.(\ref{eq:toy})).

Omitted variable bias occurs when a regression model is fitted leaving out an important causal variable. 
The problem is well known in statistics, particularly in analyzing data from experimental trials aiming at discovering causal relationships (see e.g.,  \cite{Pearl09}), but these issues have not yet been vigorously discussed in the context of  discrimination-aware data mining.

Let the true underlying model behind data be:
$y = b_0 + b_1x + \beta s  + e,$
where $x$ is a legitimate variable (such as education), $s$ is a sensitive variable (such as ethnicity), $y$ is the target variable (such as salary),  $e$ is random noise with the expected value of zero, and $\beta$, $b_1$, and $b_0$ are non-zero coefficients. 

Assume a data scientist decides to leave out the sensitive variable $s$, and fit the following model using the standard (OLS) procedure:
\begin{equation}
y = \hat{b}_0 + \hat{b}_1x.
\end{equation}

Then the estimates of the regression coefficients will be biased in the following way \cite{Zliobaite16AI}, 
\begin{eqnarray}
\hat{b}_1 & = & b_1 + \Delta,\\ \nonumber
\hat{b}_0 & = & b_0 + \beta \bar{s} - \Delta\bar{x},\\ \nonumber
\textbf{where }\Delta & = & \beta \frac{\hat{\mathit{Cov}}(x,s)}{\hat{\mathit{Var}}(x)},
\end{eqnarray}
where $\Delta$ is the bias that depends on the underlying data, $\bar{s}$ is the mean of the sensitive variable, and $\beta$ is the true underlying bias towards ethnicity in the data. 

When $s$ is omitted, $\hat{b}_1$ contains a bias, which does not go away even if we collect infinitely many observations for training the model. 
There would be no bias only in case the true $\beta = 0$, that is, the underlying data is free from inequalities in relation to the sensitive variable, or $\mathit{Cov}(x,s) = 0$, that is, the sensitive variable is not related to the legitimate variable (e.g., education variable is not related to ethnicity). 
For instance, if immigrants tend to have lower education, then the regression model would 'punish' low education even further by offering even lower wages to people with low education (who are mostly immigrants).
Thus, in most realistic cases, not only removing the sensitive variable does not make regression models fair, but on the contrary, such a strategy is likely to amplify discrimination.

We advocate that a better strategy for sanitizing regression models would be to learn a model on full data including the sensitive variable, then remove the component with the sensitive variable, and replace it by a constant that does not depend on the sensitive variable.
We advocate the following procedure and model as a baseline for non-discriminatory regression.
Firstly, build a regression model on a full dataset including the sensitive variable: 
\begin{equation}
y = b_0 + b_1x_1 + \ldots + b_kx_k + \beta s,
\label{eq:full}
\end{equation}
where $x_1,\ldots,x_k$ are legitimate input variables ($k$ is the number of variables, could be one or more), and $s$ is the sensitive variable against which discrimination is forbidden, $b_0,\ldots,b_k,\beta$ are regression coefficients.
The final model is obtained by replacing a component containing $s$ with a constant $c$:
\begin{equation}
y = b_0 + b_1x + \ldots + b_kx_k  + c,
\end{equation}
where $c$ is a constant that depends on the assumptions about the source of the underlying inequalities in the data. 
For example, one could assume that the salary paid to natives is correct, and the salary paid to immigrants is lower than it is supposed to be. 
In this case $c=0$ (assuming that $s=0$ denotes natives). 
Alternatively, one could assume that the salary paid to immigrants is correct, and natives get extra bonus. In this case $c=\beta$. 
Likely, the salary that is paid for the majority is correct, therefore, we suggest to use a weighted average of both, where the weight reflects the balance between both population groups in the data (this works for numeric sensitive variables as well, e.g., age). 
We suggest using $c = \bar{s}\beta$, where $\bar{s}$ is the mean value of the sensitive attribute over historical data, and $\beta$ is the regression coefficient from the full regression model in Eq.~(\ref{eq:full}).

\end{document}